\documentclass[10pt,twocolumn,letterpaper]{article}

\usepackage{cvpr}              
\definecolor{cvprblue}{rgb}{0.21,0.49,0.74}

\usepackage{multirow}
\usepackage{multicol}
\usepackage{amssymb}

\usepackage{pifont} 

\newcommand{\xmark}{\ding{55}}

\definecolor{emerald}{RGB}{80, 200, 120}
\definecolor{coral}{RGB}{255, 127, 80}
\definecolor{teal}{RGB}{0, 128, 128}
\definecolor{goldenrod}{RGB}{218, 165, 32}

\definecolor{darkgreen}{RGB}{0,100,0}
\definecolor{darkred}{RGB}{139,0,0}

\usepackage[pagebackref,breaklinks,colorlinks,allcolors=cvprblue]{hyperref}

\usepackage[table]{xcolor}
\usepackage{tcolorbox}

\title{Fast Reasoning Segmentation for Images and Videos}

\author{Yiqing Shen, Mathias Unberath\\
Department of Computer Science, Johns Hopkins University\\
{\tt\small \{yshen92, unberath\}@jhu.edu}
}

\begin{document}
\maketitle
\begin{abstract}
Reasoning segmentation enables open-set object segmentation via implicit text queries, therefore serving as a foundation for embodied agents that should operate autonomously in real-world environments. 
However, existing methods for reasoning segmentation require multimodal large language models (LLMs) with billions of parameters that exceed the computational capabilities of edge devices that typically deploy the embodied AI systems.
Distillation offers a pathway to compress these models while preserving their capabilities. 
Yet, existing distillation approaches fail to transfer the multi-step reasoning capabilities that reasoning segmentation demands, as they focus on matching output predictions and intermediate features rather than preserving reasoning chains.
The emerging paradigm of reasoning over digital twin representations presents an opportunity for more effective distillation by re-framing the problem.
Consequently, we propose FastReasonSeg, which employs digital twin representations that decouple perception from reasoning to enable more effective distillation.
Our distillation scheme first relies on supervised fine-tuning on teacher-generated reasoning chains.
Then it is followed by reinforcement fine-tuning with joint rewards evaluating both segmentation accuracy and reasoning quality alignment.
Experiments on two video (JiTBench, RVTBench) and two image benchmarks (ReasonSeg, LLM-Seg40K) demonstrate that our FastReasonSeg achieves state-of-the-art reasoning segmentation performance. 
Moreover, the distilled 0.6B variant outperforms models with 20$\times$ more parameters while achieving 7.79 FPS throughput with only 2.1GB memory consumption. 
This efficiency enables deployment in resource-constrained environments to enable real-time reasoning segmentation.
\end{abstract}    
\section{Introduction}
\label{sec:intro}

Reasoning segmentation enables open-set object identification through implicit text queries such as ``\textit{segment the object used to hold hot beverages}'' rather than relying on pre-defined object categories as formulated in closed-set segmentation~\cite{lisa,visa,survey}.
It contributes to more natural human-AI interactions in real-world applications such as embodied agents~\cite{miccai}, which are expected to operate in environments where the complete set of relevant objects cannot be specified in advance.
Existing reasoning segmentation methods, such as LISA~\cite{lisa} and VISA~\cite{visa}, integrate multimodal large language models (LLMs) with segmentation decoders through specialized \texttt{<SEG>} tokens that trigger mask generation. 
However, these methods demand billions of parameters, therefore creating computational and memory requirements that exceed the capabilities of edge devices where embodied agents typically operate.
Just-in-time (JiT) digital twins~\cite{rvtbench,jit,miccai} offer an alternative by constructing dynamic virtual scene representations (\textit{a}.\textit{k}.\textit{a} digital twin representation~\cite{position}) through selective activation of specialist vision models through LLM-driven planning, enabling zero-shot inference through API call of LLM. 
However, JiT's reliance on external LLM API calls introduces network latency and connectivity constraints that impede real-time performance in interactive applications. 

Training smaller reasoning segmentation models directly also faces serval limitations, as the multi-step reasoning capabilities typically manifest only when LLMs exceed certain parameter thresholds~\cite{srivastava2025towards,fu2023specializing,hsiao2025unveiling}.
Distillation offers an alternative pathway by transferring learned ``knowledge'' from large teacher models to smaller student models~\cite{liu2024contemporary,gou2021knowledge}.
However, existing distillation approaches focus on transferring output predictions and intermediate feature representations~\cite{yang2025feature}, failing to preserve the multi-step reasoning processes that enable LLMs to interpret implicit queries and navigate complex spatial-temporal relationships for reasoning segmentation~\cite{wang2024omnitokenizer}.
The architectural coupling between perception and reasoning components in current end-to-end reasoning segmentation methods exacerbates this challenge.
Visual tokenization converts continuous spatial-temporal relationships into discrete tokens, creating information bottlenecks that fragment geometric and temporal dependencies~\cite{position}.
The digital twin representation~\cite{jit,position} paradigm, however, presents an opportunity to address these distillation challenges.
By decoupling perception from reasoning through intermediate digital twin representations, they make reasoning processes explicit and transferable.
Rather than processing fragmented visual tokens, models can reason over preserved semantic, spatial, and temporal relationships organized in interpretable formats.
Such architectural separation inspire a pathway for distilling reasoning capabilities, where the structured nature of digital twin representation allows smaller models to access the same rich information as larger ones without processing high-dimensional visual inputs directly.

To address these limitations, we propose FastReasonSeg, a distillation framework that reduces computational demands by transferring reasoning capabilities from large teacher LLMs to compact student models through structured digital twin representations.
Rather than relying on visual tokenization, FastReasonSeg employs digital twin representations as structured intermediate abstractions that disentangle perception from reasoning.
It therefore enables smaller LLMs to perform complex spatial-temporal analysis without processing raw visual tokens.
The FastReasonSeg operates through a two-stage distillation process that preserves multi-step reasoning capabilities.
Initially, the teacher LLM learns to interpret implicit queries and reason over digital twin representations, refining these representations through reinforcement learning to generate accurate reasoning chains.
Then, the first distillation stage employs supervised fine-tuning, where the student LLM learns to replicate teacher-generated reasoning chains and structured output formats.
The second stage applies reinforcement learning with a reward function that evaluates both alignment to ground truth and reasoning quality relative to the teacher model, ensuring that the student maintains reasoning sophistication while operating efficiently.

The major contributions are three-fold.
First, we propose a unified reasoning segmentation framework that operates across both image and video modalities by training LLMs to perform reasoning segmentation using only digital twin representations via reinforcement learning, eliminating the need for direct visual input processing.
Second, we develop a novel two-stage distillation that preserves reasoning capabilities during knowledge transfer from large teacher LLMs to compact student models. 
It contains a supervised fine-tuning stage to establish structured reasoning patterns by learning from teacher-generated reasoning chains, followed by reinforcement learning that maintains multi-step reasoning sophistication under teacher guidance.
Third, we introduce a novel reward for the distillation training that jointly evaluates structured output format correctness and reasoning quality alignment with the teacher model.

\section{Related Works}

\paragraph{Reasoning Segmentation}
Reasoning segmentation transforms traditional object segmentation from closed-set category-based identification to open-set implicit query interpretation.
Therefore, it requires models to parse complex instructions in natural language and identify target objects through multi-step reasoning~\cite{lisa,visa,survey}.
Existing reasoning segmentation approaches follow two major paradigms, namely the end-to-end architectures or disentangled architectures that usually perform in two stages. 
End-to-end approaches like LISA~\cite{lisa}, LISA++~\cite{lisa++}, VISA~\cite{visa}, VideoLISA~\cite{videolisa}, and SegLLM~\cite{segllm} integrate multimodal LLMs with segmentation decoders using specialized tokens that trigger mask generation through supervised fine-tuning. 
Disentangled architectures such as LLM-Seg~\cite{llmseg}, SegZero~\cite{segzero}, CoReS~\cite{cores}, and JiT~\cite{jit} separate reasoning from segmentation components by preserving the capabilities of pre-trained segmentation models like SAM~\cite{sam2}.
Despite these advances, both paradigms encounter scalability challenges that limit practical deployment. 
End-to-end methods require extensive parameter counts and specialized architectural modifications, creating barriers to edge device implementation. 
Disentangled reasoning segmentation approaches, while more modular, still depend on big LLMs for reasoning, maintaining computational bottlenecks. 

\paragraph{Distillation for Model Compression}
Distillation aims to compress the model by transferring the learned representations from large teacher models to compact student models through output probability matching or feature alignment~\cite{hinton2015distilling,gou2021knowledge}. 
When applied to LLM especially the reasoning segmentation, these distillation methods typically suffer performance degradation, as compact student models struggle to replicate the intricate reasoning process~\cite{deng2023implicit,liu2024ddk}.
Consequently, reasoning-aware model compression makes an attempt to address these limitations through chain-of-thought distillation~\cite{li2024mode,deng2023implicit} and explicit multi-step reasoning transfer mechanisms~\cite{fang2025knowledge,yang2025feature}.
Nevertheless, reasoning segmentation faces additional challenges as the coupling between perception and reasoning components creates information bottlenecks.

\paragraph{Digital Twin Representation}
Digital twin representations are outcome-driven digital representations that serve as building blocks to create virtual replicas of physical processes, providing an alternative to token-based approaches that fragment continuous data in the real world into discrete tokens~\cite{position,barbie2024digital}.
They serve as structured virtual replicas that preserve semantic, spatial, and temporal relationships without direct tokenization that fragments visual information~\cite{jit,position}.
For example, just-in-time (JiT) digital twin approaches construct dynamic scene graph structures through LLM-driven planning that selectively activates vision foundation models like SAM~\cite{sam2} OWLv2~\cite{owl}, enabling LLMs to perform reasoning over these representations rather than processing raw visual tokens directly.
Hence, it can maintain the fine-grained spatial and temporal details necessary for complex reasoning segmentation tasks~\cite{jit}.
Based on JiT, recent research has used digital twin representation for automated data construction~\cite{rvtbench,shen2025temporally} and ambient monitoring~\cite{miccai}.
These approaches demonstrate that digital twin representation enables decoupling perception from reasoning. 

\begin{figure*}[!t]
\centering
\includegraphics[width=\linewidth]{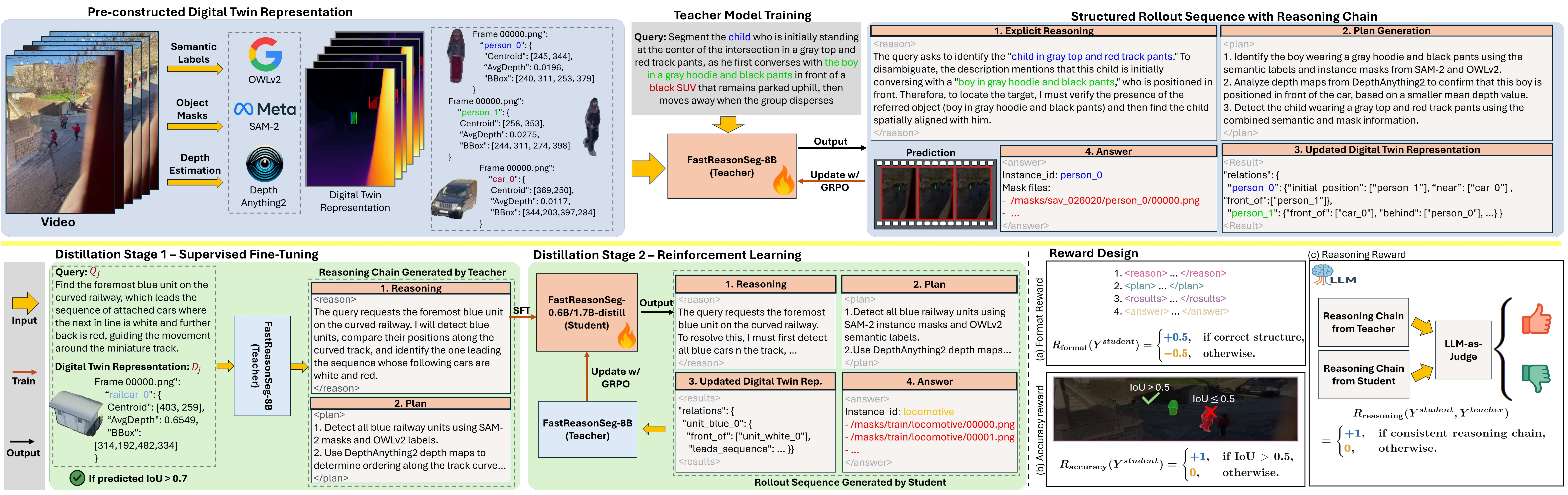}
\caption{Overall framework of the proposed FastReasonSeg.}
\label{fig:framework}
\end{figure*}

\section{Methods}

\paragraph{Overview}
As shown in Fig.~\ref{fig:framework}, the proposed FastReasonSeg aims to reduce the computational demands of reasoning segmentation models via distillation.
Rather than requiring LLMs to process raw visual tokens directly, FastReasonSeg introduces digital twin representations~\cite{position,jit} as structured intermediate abstractions that convert high-dimensional visual data into compact, semantically organized formats to enable small LLM's effective reasoning.
However, given that smaller LLMs typically cannot directly achieve generalizable reasoning capabilities through fine-tuning~\cite{srivastava2025towards}, we therefore begin by training a larger LLM first.
Specifically, this larger LLM learns to interpret implicit text queries alongside and reason over pre-constructed digital twin representations by refining the digital twin representation to generate accurate reasoning segmentation outputs.
We then transfer the knowledge in large LLM to a smaller LLM through a two-stage distillation.
The first stage employs supervised fine-tuning, where the teacher LLM generates reasoning chains from input queries, digital twin representations, and ground-truth annotations. 
The student LLM learns to replicate the structured output format and reasoning patterns demonstrated by the teacher through these generated reasoning chains via supervised fine-tuning.
The second stage applies reinforcement learning to the student's LLM using a reward function that evaluates both the alignment to ground-truth and the quality of reasoning relative to the teacher LLM.

\paragraph{Digital Twin Representation Construction}
The digital twin representation in FastReasonSeg serves as a structured intermediate layer between high-dimensional visual inputs and reasoning processes, enabling LLMs to perform reasoning without directly processing visual tokens. 
Following previous work~\cite{jit,miccai}, this representation extracts essential visual information through three complementary components that work together to capture different aspects of the visual scene.
The first component provides spatial information for individual objects through instance segmentation. 
Specifically, we employ SAM-2~\cite{sam2} to generate object masks $M^{(t)} = \{m_i^{(t)}\}_{i=1}^{N^{(t)}}$, where each $m_i^{(t)}$ represents a binary mask for the object instance $i$ at time $t$, and $N^{(t)}$ denotes the total number of detected instances in frame $t=1,\cdots,T$. 
For static images, we treat them as single-frame videos with $T=1$~\cite{sam2}. 
The second component captures three-dimensional relationships through depth estimation.
Specifically, DepthAnything2~\cite{depthanything} produces dense depth maps $Z^{(t)}$ that enable geometric relationship analysis between objects. 
For each instance, we compute depth statistics $d_i^{(t)} = \{Z^{(t)}(p) | p \in m_i^{(t)}\}$ across all pixels within the corresponding mask $m_i^{(t)}$. 
These statistics include mean depth values $\mu_i^{(t)} = \frac{1}{|m_i^{(t)}|} \sum_{p \in m_i^{(t)}} Z^{(t)}(p)$ and depth variance measures that support spatial reasoning operations such as determining relative positions and distances between objects. 
The third component provides semantic understanding through object detection and classification. 
OWLv2~\cite{owl} generates semantic labels $l_i^{(t)}$, confidence scores, and bounding box coordinates for each detected instance, allowing the reasoning process to connect visual observations with conceptual knowledge about object functions and relationships.
Formally, we define the digital twin representation at time $t$ as a structured JSON format~\cite{jit,miccai}:
\begin{equation}
\small
\begin{aligned}
D^{(t)} = \Bigg\{
i& : \Big\{
\begin{array}{l}
\text{``mask''} : m_i^{(t)}, \quad \text{``depth\_stats''} : d_i^{(t)}, \\
\text{``mean\_depth''} : \mu_i^{(t)}, \quad \text{``semantic\_label''} : l_i^{(t)}, \\
\end{array}
\Big\}\\
&\qquad \qquad \text{for } i = 1, \ldots, N^{(t)}
\Bigg\}
\end{aligned}
\end{equation}
where $l_i^{(t)}$ represents the semantic label. 
For the entire video sequence, we then can define the pre-computed digital twin representation as $\mathcal{D} = \{D^{(1)}, D^{(2)}, \ldots, D^{(T)}\}$.

\paragraph{Teacher LLM Training}
The teacher LLM learns to generate structured reasoning chains to revise the digital twin representation and resolve the reasoning segmentation by retrieving the corresponding instance masks that already existed in the digital twin representation.
Formally, given an implicit text query $Q$ and the pre-constructed digital twin representation $\mathcal{D}$, the teacher LLM produces a structured rollout sequence.
The rollout sequence begins with an explicit reasoning analysis $R$ enclosed within \texttt{<reason>} and \texttt{</reason>} tokens, where the teacher LLM examines the relationship between the query requirements and the available information within $\mathcal{D}$. 
This reasoning process determines whether the existing digital twin representation contains sufficient information to address the query or requires additional refinement to include missing contextual details.
When the LLM identifies refinement needs, it generates a revision plan $P = \{(\text{tool}_i, \text{args}_i)\}_{i=1}^{K}$ within \texttt{<plan>} and \texttt{</plan>} tokens.
Each tool-argument pair specifies a computational operation, such as instance size calculation or spatial relationship analysis, along with the corresponding parameters $\text{args}_i$ for tool calling.
Upon detecting the \texttt{</plan>} token, the atuo-regressive generation of LLM pauses to execute the specified plan, producing a refined digital twin representation $\mathcal{D}' = \{D'^{(1)}, D'^{(2)}, \ldots, D'^{(T)}\}$ that incorporates the computed enhancements.
This refined digital twin representation is then inserted within \texttt{<results>} and \texttt{</results>} tokens and appended to the rollout sequence.
The teacher LLM concludes the generation by the final segmentation answer $S$ within \texttt{<answer>} and \texttt{</answer>} tokens, identifying target objects through their corresponding mask paths and identifiers stored within the digital twin representation. 
The complete rollout sequence is formulated as follows: 
\begin{equation}
\small
Y = 
\begin{cases}
[R]_{\text{think}} \| [S]_{\text{answer}} & \text{if } P = \emptyset \\
[R]_{\text{think}} \| [P]_{\text{revise}} \| [\mathcal{D}']_{\text{results}} \| [S]_{\text{answer}} & \text{if } P \neq \emptyset
\end{cases}
\end{equation}
where $\|$ denotes token-delimited concatenation, and subscripts indicate the special token that wraps each component.

Training of the teacher LLM employs reinforcement learning with Group-Relative Policy Optimization (GRPO)~\cite{grpo} using a rule-based reward.
The total reward $\mathcal{R}(Y)$ combines two components, namely $\mathcal{R}(Y) =  \mathcal{R}_{\text{format}}(Y) + \mathcal{R}_{\text{accuracy}}(Y)$.
The format reward $\mathcal{R}_{\text{format}}(Y)$ measures the format correctness of the rollout sequence by verifying the presence and proper ordering of required token pairs, including \texttt{<reason>}, \texttt{<plan>}, \texttt{<results>}, and \texttt{<answer>} with their corresponding closing tokens.
This component returns 0.5 when the token sequence follows the correct structure and -0.5 otherwise. 
The accuracy reward $\mathcal{R}_{\text{accuracy}}(Y)$ evaluates the correctness of the final reasoning segmentation answer $S$ by comparing the identified target objects with the ground-truth annotations. 
Specifically, the correctness is determined through intersection-over-union (IoU) measurement between predicted and ground-truth masks, providing a reward of 1 when IoU exceeds 0.5 and 0 otherwise following the design in SegZero~\cite{segzero}.

\paragraph{Two-Stage Distillation}
The distillation from the teacher model to the more efficient student model proceeds through two sequential stages
The distillation process begins with supervised fine-tuning to teach student model with the output structures.
To be more specific, the teacher LLM generates the reasoning chain $\{Y_j^{\text{teacher}}\}_{j=1}^{N}$ for all training samples $\{(Q_j, \mathcal{D}_j)\}_{j=1}^{N}$.
We then use reject sampling, where we only preserve the reasoning chain if the predicted IoU is greater than 0.7.
The student model learns to replicate these teacher-generated reasoning chains through standard supervised learning objectives.
However, student LLM trained solely through supervision may struggle to adapt to scenarios that differ from the training distribution.
To address these limitations, the distillation process continues with reinforcement learning refinement that incorporates teacher guidance.
Specifically, the student LLM generates its own rollout sequences $Y^{\text{student}}$  for given query-digital twin representation pairs and receives feedback through a new reward, which enables the student LLM to maintain alignment with teacher demonstrations.
The new reward combines three components: $\mathcal{R}_{\text{total}}(Y^{\text{student}}) = \mathcal{R}_{\text{format}}(Y^{\text{student}}) + \mathcal{R}_{\text{accuracy}}(Y^{\text{student}}) + \gamma \cdot \mathcal{R}_{\text{reasoning}}(Y^{\text{student}}, Y^{\text{teacher}})$, where $\gamma$ is a balancing coefficient.
The format and accuracy rewards maintain identical definitions to those used in the training of teacher LLM.
The reasoning reward $\mathcal{R}_{\text{reasoning}}(Y^{\text{student}}, Y^{\text{teacher}})$ compares the student's reasoning chain against the teacher's reasoning chain using LLM-as-judge~\cite{llmjudging}. 
Specifically, this reward extracts the reasoning chain from both the student and teacher rollout sequences $Y^{\text{student}}$ and $Y^{\text{teacher}}$, then employs a separate LLM to assess the logical consistency, completeness, and accuracy of the student's reasoning chain relative to the teacher's in a zero-shot manner. 
The LLM judge provides scores between 0 and 1.

\section{Experiments}

\begin{table*}[!t]
\caption{Performance comparison of video reasoning segmentation on JiT benchmark~\cite{jit}. 
We report region similarity ($\mathcal{J}$) and contour accuracy ($\mathcal{F}$) across semantic, spatial, and temporal reasoning categories at three difficulty levels. 
The ``Params'' column indicates the parameter count of the backbone LLM, excluding other components. 
``API'' denotes methods that rely on external API calls to commercial LLM rather than locally deployed models.
Best results are in \textbf{bold}.
}
\label{table:jitbench}
\centering
\resizebox{\linewidth}{!}{
\begin{tabular}{l|c|ccc|c|ccc|c|ccc|c}
\toprule
\multirow{2}{*}{Methods} & \multirow{2}{*}{Params} & \multicolumn{4}{c|}{Level 1} & \multicolumn{4}{c|}{Level 2} & \multicolumn{4}{c}{Level 3} \\
\cline{3-6} \cline{7-10} \cline{11-14}
& & Semantic & Spatial & Temporal & Avg. & Semantic & Spatial & Temporal & Avg. & Semantic & Spatial & Temporal & Avg. \\
\midrule
\rowcolor{blue!15}
\multicolumn{14}{c}{\textit{Region Similarity ($\mathcal{J}$) $\uparrow$}} \\
\midrule
LISA-7B~\cite{lisa} & 7B & 0.635 & 0.226 & 0.398 & 0.420 & 0.442 & 0.213 & 0.198 & 0.284 & 0.274 & 0.229 & 0.229 & 0.244 \\
LISA-13B~\cite{lisa} & 13B & 0.669 & 0.258 & 0.237 & 0.388 & 0.472 & 0.230 & 0.176 & 0.293 & 0.301 & 0.234 & 0.177 & 0.237 \\
VISA-7B~\cite{visa} & 7B & 0.563 & 0.521 & 0.354 & 0.479 & 0.487 & 0.473 & 0.235 & 0.398 & 0.432 & 0.411 & 0.218 & 0.354 \\
VISA-13B~\cite{visa} & 13B & 0.599 & 0.532 & 0.389 & 0.507 & 0.521 & 0.505 & 0.267 & 0.431 & 0.463 & 0.441 & 0.248 & 0.384 \\
LLM-Seg~\cite{llmseg} & 7B & 0.423 & 0.315 & 0.184 & 0.307 & 0.210 & 0.201 & 0.120 & 0.177 & 0.187 & 0.154 & 0.119 & 0.153 \\
SegZero~\cite{segzero} & 3B & 0.598 & 0.445 & 0.361 & 0.468 & 0.524 & 0.398 & 0.241 & 0.388 & 0.471 & 0.352 & 0.223 & 0.349 \\
CoReS~\cite{cores} & 7B & 0.612 & 0.478 & 0.395 & 0.495 & 0.548 & 0.421 & 0.276 & 0.415 & 0.492 & 0.387 & 0.251 & 0.377 \\
CoReS~\cite{cores} & 13B & 0.628 & 0.492 & 0.408 & 0.509 & 0.564 & 0.435 & 0.289 & 0.429 & 0.508 & 0.401 & 0.264 & 0.391 \\
JiT~\cite{jit} & 7B & 0.634 & 0.495 & 0.412 & 0.514 & 0.572 & 0.448 & 0.302 & 0.441 & 0.516 & 0.418 & 0.275 & 0.403 \\
JiT~\cite{jit} & API & 0.865 & 0.789 & 0.721 & 0.792 & 0.841 & 0.752 & 0.705 & 0.766 & 0.810 & 0.741 & 0.690 & 0.747 \\
\midrule
\rowcolor{gray!10}
FastReasonSeg-8B & 8B & \textbf{0.882} & \textbf{0.807} & \textbf{0.738} & \textbf{0.809} & \textbf{0.857} & \textbf{0.769} & \textbf{0.719} & \textbf{0.782} & \textbf{0.824} & \textbf{0.756} & \textbf{0.703} & \textbf{0.761} \\
\rowcolor{gray!10}
FastReasonSeg-1.7B-Distill & 1.7B & 0.856 & 0.781 & 0.715 & 0.784 & 0.832 & 0.744 & 0.697 & 0.758 & 0.801 & 0.732 & 0.681 & 0.738 \\
\rowcolor{gray!10}
FastReasonSeg-0.6B-Distill & 0.6B & 0.831 & 0.756 & 0.692 & 0.760 & 0.807 & 0.719 & 0.674 & 0.733 & 0.776 & 0.707 & 0.658 & 0.714 \\
\midrule
\rowcolor{blue!15}
\multicolumn{14}{c}{\textit{Contour Accuracy ($\mathcal{F}$) $\uparrow$}} \\
\midrule
LISA-7B~\cite{lisa} & 7B & 0.706 & 0.283 & 0.451 & 0.480 & 0.490 & 0.268 & 0.273 & 0.344 & 0.322 & 0.282 & 0.307 & 0.304 \\
LISA-13B~\cite{lisa} & 13B & 0.756 & 0.313 & 0.320 & 0.463 & 0.524 & 0.283 & 0.256 & 0.354 & 0.353 & 0.280 & 0.259 & 0.297 \\
VISA-7B~\cite{visa} & 7B & 0.585 & 0.563 & 0.327 & 0.492 & 0.514 & 0.510 & 0.303 & 0.442 & 0.497 & 0.499 & 0.277 & 0.424 \\
VISA-13B~\cite{visa} & 13B & 0.621 & 0.587 & 0.381 & 0.530 & 0.547 & 0.543 & 0.334 & 0.475 & 0.528 & 0.531 & 0.306 & 0.455 \\
LLM-Seg~\cite{llmseg} & 7B & 0.535 & 0.345 & 0.278 & 0.386 & 0.437 & 0.258 & 0.247 & 0.314 & 0.319 & 0.218 & 0.218 & 0.252 \\
SegZero~\cite{segzero} & 3B & 0.622 & 0.478 & 0.334 & 0.478 & 0.548 & 0.432 & 0.309 & 0.430 & 0.523 & 0.421 & 0.281 & 0.408 \\
CoReS~\cite{cores} & 7B & 0.638 & 0.492 & 0.368 & 0.499 & 0.562 & 0.445 & 0.327 & 0.445 & 0.536 & 0.438 & 0.294 & 0.423 \\
CoReS~\cite{cores} & 13B & 0.654 & 0.506 & 0.382 & 0.514 & 0.578 & 0.459 & 0.341 & 0.459 & 0.552 & 0.452 & 0.308 & 0.437 \\
JiT~\cite{jit} & 7B & 0.661 & 0.518 & 0.396 & 0.525 & 0.589 & 0.472 & 0.354 & 0.472 & 0.564 & 0.465 & 0.319 & 0.449 \\
JiT~\cite{jit} & API & 0.795 & 0.831 & 0.793 & 0.806 & 0.801 & 0.819 & 0.784 & 0.801 & 0.801 & 0.792 & 0.737 & 0.777 \\
\midrule
\rowcolor{gray!10}
FastReasonSeg-8B & 8B & \textbf{0.808} & \textbf{0.844} & \textbf{0.806} & \textbf{0.819} & \textbf{0.814} & \textbf{0.831} & \textbf{0.797} & \textbf{0.814} & \textbf{0.813} & \textbf{0.804} & \textbf{0.748} & \textbf{0.788} \\
\rowcolor{gray!10}
FastReasonSeg-1.7B-Distill & 1.7B & 0.782 & 0.818 & 0.780 & 0.793 & 0.788 & 0.805 & 0.772 & 0.788 & 0.788 & 0.779 & 0.724 & 0.764 \\
\rowcolor{gray!10}
FastReasonSeg-0.6B-Distill & 0.6B & 0.757 & 0.793 & 0.755 & 0.768 & 0.763 & 0.780 & 0.747 & 0.763 & 0.763 & 0.754 & 0.699 & 0.739 \\
\bottomrule
\end{tabular}
}
\end{table*}

\begin{table*}[!t]
\caption{Performance comparison of video reasoning segmentation methods on RVTBench~\cite{rvtbench} segmentation subset. 
%
}
\label{table:rvtbench_seg}
\centering
\resizebox{\linewidth}{!}{
\begin{tabular}{l|c|ccc|c|ccc|c|ccc|c|ccc|c}
\toprule
\multirow{2}{*}{Methods} & \multirow{2}{*}{Params} & \multicolumn{4}{c|}{Level 1} & \multicolumn{4}{c|}{Level 2} & \multicolumn{4}{c|}{Level 3} & \multicolumn{4}{c}{Level 4} \\
\cline{3-6} \cline{7-10} \cline{11-14} \cline{15-18}
& & Semantic & Spatial & Temporal & Avg. & Semantic & Spatial & Temporal & Avg. & Semantic & Spatial & Temporal & Avg. & Semantic & Spatial & Temporal & Avg. \\
\midrule
\rowcolor{blue!15}
\multicolumn{18}{c}{\textit{Region Similarity ($\mathcal{J}$) $\uparrow$}} \\
\midrule
LISA-7B~\cite{lisa} & 7B & 0.458 & 0.432 & 0.416 & 0.435 & 0.449 & 0.421 & 0.390 & 0.420 & 0.421 & 0.406 & 0.366 & 0.398 & 0.404 & 0.383 & 0.343 & 0.377 \\
LISA-13B~\cite{lisa} & 13B & 0.485 & 0.465 & 0.448 & 0.466 & 0.478 & 0.457 & 0.421 & 0.452 & 0.458 & 0.444 & 0.391 & 0.431 & 0.439 & 0.426 & 0.368 & 0.411 \\
VISA-7B~\cite{visa} & 7B & 0.453 & 0.439 & 0.395 & 0.429 & 0.440 & 0.422 & 0.373 & 0.412 & 0.425 & 0.406 & 0.352 & 0.394 & 0.401 & 0.383 & 0.331 & 0.372 \\
VISA-13B~\cite{visa} & 13B & 0.512 & 0.488 & 0.444 & 0.481 & 0.495 & 0.472 & 0.419 & 0.462 & 0.477 & 0.456 & 0.396 & 0.443 & 0.452 & 0.431 & 0.372 & 0.418 \\
LLM-Seg~\cite{llmseg} & 7B & 0.095 & 0.096 & 0.200 & 0.130 & 0.095 & 0.092 & 0.129 & 0.105 & 0.096 & 0.094 & 0.117 & 0.103 & 0.094 & 0.091 & 0.106 & 0.097 \\
SegZero~\cite{segzero} & 3B & 0.479 & 0.446 & 0.413 & 0.446 & 0.459 & 0.424 & 0.387 & 0.423 & 0.438 & 0.408 & 0.365 & 0.404 & 0.413 & 0.389 & 0.347 & 0.383 \\
CoReS~\cite{cores} & 7B & 0.467 & 0.442 & 0.401 & 0.437 & 0.451 & 0.425 & 0.382 & 0.419 & 0.434 & 0.408 & 0.361 & 0.401 & 0.409 & 0.385 & 0.342 & 0.379 \\
CoReS~\cite{cores} & 13B & 0.481 & 0.456 & 0.415 & 0.451 & 0.465 & 0.439 & 0.396 & 0.433 & 0.448 & 0.422 & 0.375 & 0.415 & 0.423 & 0.399 & 0.356 & 0.393 \\
JiT~\cite{jit} & 7B & 0.421 & 0.398 & 0.365 & 0.395 & 0.406 & 0.382 & 0.344 & 0.377 & 0.392 & 0.367 & 0.325 & 0.361 & 0.371 & 0.348 & 0.307 & 0.342 \\
JiT~\cite{jit} & API & 0.548 & 0.521 & 0.475 & 0.515 & 0.529 & 0.504 & 0.448 & 0.494 & 0.510 & 0.487 & 0.423 & 0.474 & 0.483 & 0.461 & 0.399 & 0.448 \\
\midrule
\rowcolor{gray!10}
FastReasonSeg-8B & 8B & \textbf{0.715} & \textbf{0.682} & \textbf{0.638} & \textbf{0.678} & \textbf{0.688} & \textbf{0.657} & \textbf{0.604} & \textbf{0.650} & \textbf{0.663} & \textbf{0.632} & \textbf{0.577} & \textbf{0.624} & \textbf{0.638} & \textbf{0.608} & \textbf{0.552} & \textbf{0.599} \\
\rowcolor{gray!10}
FastReasonSeg-1.7B-Distill & 1.7B & 0.702 & 0.669 & 0.625 & 0.665 & 0.675 & 0.644 & 0.591 & 0.637 & 0.650 & 0.619 & 0.564 & 0.611 & 0.625 & 0.595 & 0.539 & 0.586 \\
\rowcolor{gray!10}
FastReasonSeg-0.6B-Distill & 0.6B & 0.685 & 0.652 & 0.609 & 0.649 & 0.658 & 0.627 & 0.575 & 0.620 & 0.633 & 0.602 & 0.548 & 0.594 & 0.608 & 0.578 & 0.523 & 0.570 \\
\midrule
\rowcolor{blue!15}
\multicolumn{18}{c}{\textit{Contour Accuracy ($\mathcal{F}$) $\uparrow$}} \\
\midrule
LISA-7B~\cite{lisa} & 7B & 0.422 & 0.407 & 0.394 & 0.408 & 0.412 & 0.396 & 0.357 & 0.388 & 0.394 & 0.381 & 0.332 & 0.369 & 0.366 & 0.355 & 0.310 & 0.343 \\
LISA-13B~\cite{lisa} & 13B & 0.459 & 0.447 & 0.425 & 0.444 & 0.452 & 0.433 & 0.394 & 0.426 & 0.431 & 0.418 & 0.367 & 0.405 & 0.413 & 0.401 & 0.346 & 0.387 \\
VISA-7B~\cite{visa} & 7B & 0.437 & 0.422 & 0.380 & 0.413 & 0.425 & 0.407 & 0.356 & 0.396 & 0.410 & 0.393 & 0.336 & 0.379 & 0.387 & 0.371 & 0.315 & 0.358 \\
VISA-13B~\cite{visa} & 13B & 0.491 & 0.473 & 0.426 & 0.463 & 0.477 & 0.458 & 0.401 & 0.445 & 0.461 & 0.442 & 0.378 & 0.427 & 0.436 & 0.417 & 0.354 & 0.402 \\
LLM-Seg~\cite{llmseg} & 7B & 0.130 & 0.129 & 0.244 & 0.168 & 0.132 & 0.129 & 0.168 & 0.143 & 0.132 & 0.130 & 0.157 & 0.140 & 0.129 & 0.127 & 0.145 & 0.134 \\
SegZero~\cite{segzero} & 3B & 0.461 & 0.437 & 0.399 & 0.432 & 0.443 & 0.416 & 0.372 & 0.410 & 0.422 & 0.398 & 0.349 & 0.390 & 0.398 & 0.375 & 0.328 & 0.367 \\
CoReS~\cite{cores} & 7B & 0.452 & 0.428 & 0.387 & 0.422 & 0.436 & 0.412 & 0.368 & 0.405 & 0.419 & 0.395 & 0.346 & 0.387 & 0.395 & 0.372 & 0.325 & 0.364 \\
CoReS~\cite{cores} & 13B & 0.466 & 0.442 & 0.401 & 0.436 & 0.450 & 0.426 & 0.382 & 0.419 & 0.433 & 0.409 & 0.360 & 0.401 & 0.409 & 0.386 & 0.339 & 0.378 \\
JiT~\cite{jit} & 7B & 0.407 & 0.385 & 0.352 & 0.381 & 0.393 & 0.370 & 0.331 & 0.365 & 0.379 & 0.356 & 0.311 & 0.349 & 0.359 & 0.337 & 0.292 & 0.329 \\
JiT~\cite{jit} & API & 0.526 & 0.506 & 0.457 & 0.496 & 0.510 & 0.490 & 0.429 & 0.476 & 0.493 & 0.472 & 0.404 & 0.456 & 0.467 & 0.447 & 0.379 & 0.431 \\
\midrule
\rowcolor{gray!10}
FastReasonSeg-8B & 8B & \textbf{0.692} & \textbf{0.669} & \textbf{0.616} & \textbf{0.659} & \textbf{0.665} & \textbf{0.643} & \textbf{0.582} & \textbf{0.630} & \textbf{0.639} & \textbf{0.619} & \textbf{0.558} & \textbf{0.605} & \textbf{0.612} & \textbf{0.591} & \textbf{0.534} & \textbf{0.579} \\
\rowcolor{gray!10}
FastReasonSeg-1.7B-Distill & 1.7B & 0.679 & 0.656 & 0.603 & 0.646 & 0.652 & 0.630 & 0.569 & 0.617 & 0.626 & 0.606 & 0.545 & 0.592 & 0.599 & 0.578 & 0.521 & 0.566 \\
\rowcolor{gray!10}
FastReasonSeg-0.6B-Distill & 0.6B & 0.662 & 0.639 & 0.587 & 0.629 & 0.635 & 0.613 & 0.553 & 0.600 & 0.609 & 0.589 & 0.529 & 0.576 & 0.582 & 0.561 & 0.505 & 0.549 \\
\bottomrule
\end{tabular}
}
\end{table*}

\paragraph{Implementation Details}

We implement the teacher model in FastReasonSeg with Qwen3-8B~\cite{qwen3}; while the student models utilize Qwen3-1.7B and Qwen3-0.6B variants.
Training is carried out using LoRA~\cite{lora} on 8 NVIDIA GeForce RTX 4090 GPUs, gradient accumulation with a step size of 64 to achieve an effective batch size of 512. 
The LoRA configuration uses rank $r = 64$ with $\alpha = 128$.
Mixed precision training utilizes automatic FP16 scaling to optimize memory utilization on the 24GB GPU memory constraint.
The teacher model training employs GRPO~\cite{grpo} with learning rate $4 \times 10^{-5}$, linear warm-up over 10\% of training steps, and cosine annealing scheduler.
In distillation, the supervised fine-tuning stage utilizes AdamW optimizer with learning rate $5 \times 10^{-5}$, weight decay of $0.01$, and trains for 3 epochs.
The reinforcement learning distillation stage employs the same GRPO configuration as teacher training but incorporates the reasoning quality reward with coefficient $\gamma = 0.5$. 
For the LLM-as-judge reward component, we use GPT-4o~\cite{gpt4o} with temperature $0.3$ and a maximum token limit of 512 for reasoning chain evaluation. 
For training, we use RefCOCOg~\cite{refcocog} and training sets of ReasonSeg~\cite{lisa}, and ReVOS~\cite{visa}.

\paragraph{Datasets and Evaluation Metrics}
We evaluate FastReasonSeg across benchmarks spanning both image and video reasoning segmentation tasks. 
For video reasoning segmentation, we conduct experiments on JiTBench~\cite{jit}, which contains 895 queries across three reasoning categories with three difficulty levels.
We also use the RVTBench~\cite{rvtbench} reasoning segmentation subset, which comprises three reasoning categories and four difficulty levels derived from 200 video sequences. 
For image reasoning segmentation, we leverage the test sets from ReasonSeg~\cite{lisa}, containing both short and long query formats to evaluate comprehension of implicit reasoning instructions, and LLM-Seg40K~\cite{llmseg} to demonstrate cross-modality generalization capabilities.
Video reasoning segmentation performance is measured using region similarity ($\mathcal{J}$) and contour accuracy ($\mathcal{F}$)~\cite{davis}.
Image reasoning segmentation adopts cIoU and gIoU as the metrics~\cite{lisa}.

\begin{table*}[!t]
\caption{Ablation study on JiTBench~\cite{jit} evaluating the contribution of each component in FastReasonSeg-1.7B-Distill.
We report both region similarity ($\mathcal{J}$) and contour accuracy ($\mathcal{F}$) averaged across all reasoning categories for each difficulty level. 
``DT'' denotes digital twin representation, ``SFT'' denotes supervised fine-tuning stage, and ``RL'' denotes reinforcement learning stage. 
The asterisk (*) indicates using static digital twin without dynamic refinement through tool calling.}
\label{table:ablation}
\centering
\resizebox{\linewidth}{!}{
\begin{tabular}{l|ccccc|ccc|c|ccc|c}
\toprule
\multirow{3}{*}{Configuration} & \multicolumn{5}{c|}{Components} & \multicolumn{8}{c}{Performance Metrics} \\
\cline{2-14}
& \multirow{2}{*}{DT} & \multirow{2}{*}{SFT} & \multirow{2}{*}{RL} & \multirow{2}{*}{Reasoning} & \multirow{2}{*}{Format} & \multicolumn{4}{c|}{Region Similarity ($\mathcal{J}$)} & \multicolumn{4}{c}{Contour Accuracy ($\mathcal{F}$)} \\
\cline{7-10} \cline{11-14}
& Rep. & Stage & Stage & Reward & Reward & Level 1 & Level 2 & Level 3 & Avg. & Level 1 & Level 2 & Level 3 & Avg. \\
\midrule
\rowcolor{gray!10}
Full Model (FastReasonSeg-1.7B-Distill) & \textcolor{darkgreen}{\checkmark} & \textcolor{darkgreen}{\checkmark} & \textcolor{darkgreen}{\checkmark} & \textcolor{darkgreen}{\checkmark} & \textcolor{darkgreen}{\checkmark} & 0.784 & 0.758 & 0.738 & 0.760 & 0.793 & 0.788 & 0.764 & 0.782 \\
\midrule
\rowcolor{blue!15}
\multicolumn{14}{l}{\textit{Representation Alternatives}} \\
w/o DT (Direct Visual Tokens) & \textcolor{darkred}{\xmark} & \textcolor{darkgreen}{\checkmark} & \textcolor{darkgreen}{\checkmark} & \textcolor{darkgreen}{\checkmark} & \textcolor{darkgreen}{\checkmark} & 0.612 & 0.587 & 0.564 & 0.588 & 0.625 & 0.608 & 0.591 & 0.608 \\
w/o DT Refinement & \textcolor{darkgreen}{\checkmark}$^*$ & \textcolor{darkgreen}{\checkmark} & \textcolor{darkgreen}{\checkmark} & \textcolor{darkgreen}{\checkmark} & \textcolor{darkgreen}{\checkmark} & 0.723 & 0.698 & 0.675 & 0.699 & 0.734 & 0.721 & 0.698 & 0.718 \\
\midrule
\rowcolor{blue!15}
\multicolumn{14}{l}{\textit{Training Stage Ablations}} \\
w/o Two-Stage (Direct RL Only) & \textcolor{darkgreen}{\checkmark} & \textcolor{darkred}{\xmark} & \textcolor{darkgreen}{\checkmark} & \textcolor{darkgreen}{\checkmark} & \textcolor{darkgreen}{\checkmark} & 0.695 & 0.668 & 0.643 & 0.669 & 0.706 & 0.691 & 0.667 & 0.688 \\
w/o RL (SFT Only) & \textcolor{darkgreen}{\checkmark} & \textcolor{darkgreen}{\checkmark} & \textcolor{darkred}{\xmark} & \textcolor{darkred}{\xmark} & \textcolor{darkred}{\xmark} & 0.742 & 0.715 & 0.691 & 0.716 & 0.751 & 0.738 & 0.713 & 0.734 \\
\midrule
\rowcolor{blue!15}
\multicolumn{14}{l}{\textit{Reward Function Components}} \\
w/o Reasoning Reward & \textcolor{darkgreen}{\checkmark} & \textcolor{darkgreen}{\checkmark} & \textcolor{darkgreen}{\checkmark} & \textcolor{darkred}{\xmark} & \textcolor{darkgreen}{\checkmark} & 0.761 & 0.734 & 0.713 & 0.736 & 0.770 & 0.759 & 0.738 & 0.756 \\
w/o Format Reward & \textcolor{darkgreen}{\checkmark} & \textcolor{darkgreen}{\checkmark} & \textcolor{darkgreen}{\checkmark} & \textcolor{darkgreen}{\checkmark} & \textcolor{darkred}{\xmark} & 0.718 & 0.689 & 0.665 & 0.691 & 0.728 & 0.712 & 0.689 & 0.710 \\
w/o Teacher Guidance & \textcolor{darkgreen}{\checkmark} & \textcolor{darkgreen}{\checkmark} & \textcolor{darkgreen}{\checkmark} & \textcolor{darkred}{\xmark} & \textcolor{darkred}{\xmark} & 0.704 & 0.676 & 0.651 & 0.677 & 0.714 & 0.699 & 0.674 & 0.696 \\
\midrule
\rowcolor{blue!15}
\multicolumn{14}{l}{\textit{Teacher Model Variations}} \\
Teacher w/ Smaller LLM (3B) & \textcolor{darkgreen}{\checkmark} & \textcolor{darkgreen}{\checkmark} & \textcolor{darkgreen}{\checkmark} & \textcolor{darkgreen}{\checkmark} & \textcolor{darkgreen}{\checkmark} & 0.728 & 0.701 & 0.678 & 0.702 & 0.738 & 0.724 & 0.701 & 0.721 \\
Teacher w/o RL Training & \textcolor{darkgreen}{\checkmark} & \textcolor{darkgreen}{\checkmark} & \textcolor{darkgreen}{\checkmark} & \textcolor{darkgreen}{\checkmark} & \textcolor{darkgreen}{\checkmark} & 0.735 & 0.708 & 0.684 & 0.709 & 0.745 & 0.731 & 0.707 & 0.728 \\
\bottomrule
\end{tabular}
}
\end{table*}

\paragraph{Compared Methods}
We compare FastReasonSeg with state-of-the-art reasoning segmentation methods. 
For segmentation of image reasoning, LISA~\cite{lisa} is an end-to-end approach by extending multimodal LLMs (LLaVA-7B/13B-v1-1~\cite{llava}) with a specialized \texttt{<SEG>} token and embedding-as-mask paradigm to generate segmentation masks from implicit queries. 
CoReS~\cite{cores} improves LISA through a design of dual chain of thought~\cite{cot} built on LLaVA-7B/13B-v0~\cite{llava} that mimics visual search for human cognition.
LLM-Seg~\cite{llmseg} proposes a two-stage image reasoning segmentation approach connecting LLMs (LLaVA-lightning-7B~\cite{llava}) with SAM~\cite{sam1} through mask proposal selection rather than direct token generation. 
Similarly, SegZero~\cite{segzero} introduces a decoupled image reasoning segmentation architecture with an LLM (Qwen2.5-VL-3B~\cite{qwenvl}) and a segmentation model (SAM~\cite{sam1}), trained through reinforcement learning without explicit reasoning data.
To evaluate these image-based methods on video reasoning segmentation benchmarks, we follow previous works~\cite{visa,jit} by adapting them to process video sequences through frame-by-frame inference. 
The comparison includes VISA~\cite{visa} which uses Chat-UniVi-7B/13B~\cite{chatuniv} as the LLM backbone by incorporating text-guided frame sampling and multimodal LLM reasoning.  
We also compare with JiT~\cite{jit}, which represents the agent-based paradigm using digital twin representations for video reasoning segmentation without requiring LLM fine-tuning. 
JiT employs GPT-4o for API-based inference and locally deployed DeepSeek-R1-Distill-Qwen-7B~\cite{r1} for 7B model configurations.

\begin{figure}[!ht]
\centering
\includegraphics[width=\linewidth]{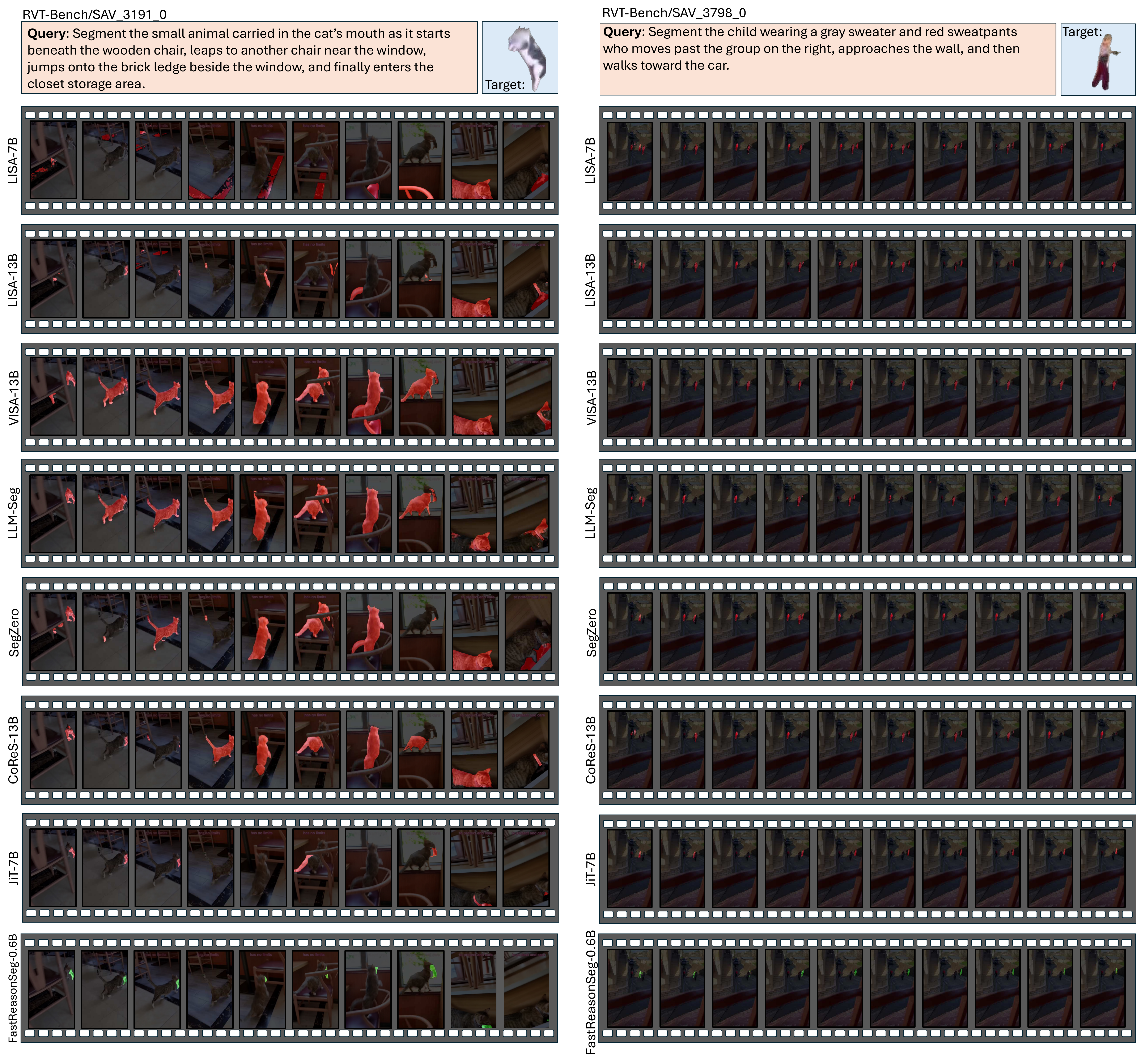}
\caption{Qualitative comparison of video reasoning segmentation results on JiTBench. 
The figure displays two representative examples requiring complex spatial-temporal reasoning.
%
%
Each row presents segmentation outputs from different methods across video frames, with green masks indicating accurate predictions and red masks denoting incorrect segmentations. 
FastReasonSeg-0.6B produces more accurate and consistent masks across temporal sequences compared to baseline approaches, despite operating with fewer parameters than competing methods}
\label{fig:result}
\end{figure}

\paragraph{Performance for Video Reasoning Segmentation}
Our evaluation in Tables~\ref{table:jitbench} and~\ref{table:rvtbench_seg} and Fig.~\ref{fig:result} on video reasoning segmentation benchmarks demonstrates the effectiveness of the proposed distillation framework in multiple reasoning categories and difficulty levels. 
In JiTBench~\cite{jit}, FastReasonSeg-8B achieves the highest performance across all reasoning categories and difficulty levels. 
For region similarity ($\mathcal{J}$), our FastReasonSeg-8B teacher model achieves average scores of 0.809, 0.782, and 0.761 for levels 1, 2, and 3, respectively, surpassing the strongest baseline JiT with GPT4o by margins of 0.017, 0.016, and 0.014. 
Performance gains are particularly pronounced in spatial reasoning tasks, where FastReasonSeg-8B achieves 0.807 at level 1 compared to 0.789 with JiT with GPT4o. 
Similar patterns emerge in contour accuracy ($\mathcal{F}$), with FastReasonSeg-8B achieving consistent improvements in all scenarios.
Importantly, FastReasonSeg-1.7B-Distill maintains competitive performance with average region similarity scores of 0.784, 0.758, and 0.738 across the three difficulty levels. 
Even the most compact FastReasonSeg-0.6B-Distill variant outperforms established methods such as VISA-13B~\cite{visa} and CoReS-13B~\cite{cores}, which contain more than 20 times more parameters.
The distilled variants continue to demonstrate superior performance on RVTBench~\cite{rvtbench}. 
FastReasonSeg-1.7B-Distill achieves region similarity scores that exceed all baseline methods at all difficulty levels, while FastReasonSeg-0.6B-Distill maintains competitive performance despite operating with minimal computational requirements.
%

\begin{table}[!ht]
\caption{
Performance evaluation of image reasoning segmentation on ReasonSeg~\cite{lisa} and LLM-Seg40K~\cite{llmseg} via gIoU and cIoU.
}
\label{table:image_rs}
\centering
\resizebox{\linewidth}{!}{
\begin{tabular}{l|c|cc|cc|cc} 
\toprule
\multirow{3}{*}{Methods} & \multirow{3}{*}{Params} & \multicolumn{4}{c|}{ReasonSeg~\cite{lisa}} & \multicolumn{2}{c}{LLM-Seg40K~\cite{llmseg}} \\
\cline{3-6} \cline{7-8}
& & \multicolumn{2}{c|}{Short Query} & \multicolumn{2}{c|}{Long Query} & \multicolumn{2}{c}{Overall} \\
\cline{3-4} \cline{5-6} \cline{7-8}
& & gIoU & cIoU & gIoU & cIoU & gIoU & cIoU \\
\midrule
LISA-7B~\cite{lisa} & 7B & 0.483 & 0.463 & 0.579 & 0.597 & 0.376 & 0.485 \\
LISA-13B~\cite{lisa} & 13B & 0.554 & 0.506 & 0.632 & 0.653 & 0.392 & 0.502 \\
VISA-7B~\cite{visa} & 7B & 0.453 & 0.482 & 0.453 & 0.455 & 0.358 & 0.394 \\
VISA-13B~\cite{visa} & 13B & 0.497 & 0.521 & 0.498 & 0.502 & 0.381 & 0.417 \\
LLM-Seg~\cite{llmseg} & 7B & 0.210 & 0.203 & 0.253 & 0.248 & 0.455 & 0.542 \\
Seg-Zero~\cite{segzero} & 3B & 0.531 & 0.498 & 0.587 & 0.612 & 0.442 & 0.478 \\
CoReS~\cite{cores} & 7B & 0.548 & 0.515 & 0.603 & 0.628 & 0.458 & 0.495 \\
CoReS~\cite{cores} & 13B & 0.565 & 0.532 & 0.621 & 0.645 & 0.474 & 0.511 \\
JiT~\cite{jit} & 7B & 0.521 & 0.487 & 0.579 & 0.601 & 0.398 & 0.435 \\
JiT~\cite{jit} & API & 0.618 & 0.584 & 0.683 & 0.701 & 0.485 & 0.528 \\
\midrule
\rowcolor{gray!10}
FastReasonSeg-8B & 8B & \textbf{0.746} & \textbf{0.713} & \textbf{0.812} & \textbf{0.834} & \textbf{0.641} & \textbf{0.677} \\
\rowcolor{gray!10}
FastReasonSeg-1.7B-Distill & 1.7B & 0.721 & 0.688 & 0.787 & 0.809 & 0.618 & 0.654 \\
\rowcolor{gray!10}
FastReasonSeg-0.6B-Distill & 0.6B & 0.696 & 0.663 & 0.762 & 0.784 & 0.595 & 0.631 \\
\bottomrule
\end{tabular}
}
\end{table}

\paragraph{Performance for Image Reasoning Segmentation}
Table~\ref{table:image_rs} presents the evaluation results on image reasoning segmentation benchmarks, showcasing the cross-modality generalization capabilities of FastReasonSeg. 
On the ReasonSeg benchmark, FastReasonSeg-8B establishes new state-of-the-art results across both short and long query categories. 
For short queries, our teacher model reaches 0.746 gIoU and 0.713 cIoU, representing improvements of 0.128 and 0.129 over the previous best method JiT with GPT4o. 
The performance advantage becomes more observable with long queries, where FastReasonSeg-8B achieves 0.812 gIoU and 0.834 cIoU, exceeding JiT with GPT4o by margins of 0.129 and 0.133, respectively. 
The LLM-Seg40K evaluation reveals compelling evidence of our approach's robustness across different dataset characteristics. 
FastReasonSeg-8B achieves 0.641 gIoU and 0.677 cIoU, substantially outperforming existing methods. 
Notably, while LLM-Seg shows competitive cIoU performance (0.542) on its native benchmark, our approach delivers superior results across both metrics, demonstrating the broader applicability of our digital twin-based reasoning framework.

FastReasonSeg-1.7B-Distill delivers 0.721 and 0.787 gIoU scores for short and long queries, respectively, while operating with approximately one-fifth the parameters of competing 7B models. 
This represents an improvement in computational efficiency without sacrificing reasoning quality. 
The compact FastReasonSeg-0.6B-Distill variant further demonstrates the scalability of our distillation approach, achieving 0.696 gIoU in short queries while requiring only 0.6B parameters.
These results of image reasoning segmentation complement our video evaluation findings and demonstrate the unified nature of our method across different modalities.

\begin{table}[!ht]
\caption{Efficiency comparison of reasoning segmentation methods in terms of computational requirements and runtime characteristics.
``Total Params'' indicates the complete model parameters including segmentation components.
``LLM Params'' represents parameters of the LLM backbone only. 
``Memory'' represents peak GPU memory consumption during inference. 
``FLOPs'' measures computational operations per frame. ``Latency'' reports average processing time per frame. ``Throughput'' indicates frames processed per second. 
For JiT and our method, ``Total Params'', ``Memory'' and ``FLOPs'' also include vision foundation models used for DT representation construction.
All metrics are computed on averaged.
Best efficiency metrics are highlighted in \textbf{bold}.}
\label{table:efficiency_comparison}
\centering
\resizebox{\linewidth}{!}{
\begin{tabular}{l|c|c|c|c|c|c}
\toprule
Methods & \begin{tabular}{c}Total\\Params (B)\end{tabular} & \begin{tabular}{c}LLM\\Params (B)\end{tabular} & \begin{tabular}{c}Memory\\(GB)\end{tabular} & \begin{tabular}{c}FLOPs\\(G)\end{tabular} & \begin{tabular}{c}Latency\\(ms)\end{tabular} & \begin{tabular}{c}Throughput\\(FPS)\end{tabular} \\
\midrule
LISA-7B~\cite{lisa} & 7.7 & 7.0 & 16.0 & 24.2 & 749.7 & 1.33 \\
LISA-13B~\cite{lisa} & 14.0 & 13.0 & 27.88 & 43.7 & 1128.9 & 0.89 \\
VISA-7B~\cite{visa} & 7.7 & 7.0 & 15.4 & 28.7 & 534.0 & 1.87 \\
VISA-13B~\cite{visa} & 14.0 & 13.0 & 28.0 & 52.1 & 629.8 & 1.59 \\
LLM-Seg~\cite{llmseg} & 8.014 & 7.0 & 16.66 & 37.57 & 3469.6 & 0.29 \\
SegZero~\cite{segzero} & 8.5 & 3.0 & 17.03 & 10.3 & 9302.3 & 0.11 \\
CoReS~\cite{cores} & 7.7 & 7.0 & 16.1 & 88.6 & 3762.3 & 0.27 \\
CoReS~\cite{cores} & 14.0 & 13.0 & 28.32 & 174.8 & 6391.9 & 0.16 \\
JiT~\cite{jit} & 8.6 & 7.0 & 17.2 & 21.7 & 39914.7 & 0.03 \\
JiT~\cite{jit} & - & API & - & - & 52845.7 & 0.02 \\
\midrule
\rowcolor{gray!10}
FastReasonSeg-8B & 8.2 & 8.0 & 18.5 & 49.2 & 892.5 & 1.12 \\
\rowcolor{gray!10}
FastReasonSeg-1.7B-Distill & 1.9 & 1.7 & 4.2 & 12.3 & 245.7 & 4.07 \\
\rowcolor{gray!10}
FastReasonSeg-0.6B-Distill & \textbf{0.8} & \textbf{0.6} & \textbf{2.1} & \textbf{6.8} & \textbf{128.4} & \textbf{7.79} \\
\bottomrule
\end{tabular}
}
\end{table}

\paragraph{Comparison of Model Efficiency}

Table~\ref{table:efficiency_comparison} presents the computational efficiency analysis.
FastReasonSeg-0.6B-Distill achieves the best efficiency metrics with only 0.8B total parameters, 2.1GB memory consumption, 6.8G FLOPs per frame, 128.4ms latency, and 7.79 FPS throughput. 
The distilled models operate with considerably lower computational overhead compared to traditional approaches, where methods like LISA-13B~\cite{lisa} require 14.0B parameters and 28.0GB memory while achieving only 0.89 FPS. 
Agent-based approaches such as JiT~\cite{jit} with 7B LLM backbone suffer from extremely high latency, making them impractical for real-time applications. 
Our FastReasonSeg-1.7B-Distill model delivers 4.07 FPS with 1.9B parameters, representing more than a 4$\times$ speedup over the fastest baseline VISA-7B while maintaining better performance. 
The efficiency gains become more observable when comparing against methods with similar reasoning complexity, for example CoReS-13B requires 6,391.9ms latency compared to our teacher model's 892.5ms.

\begin{figure}[!ht]
\centering
\includegraphics[width=0.95\linewidth]{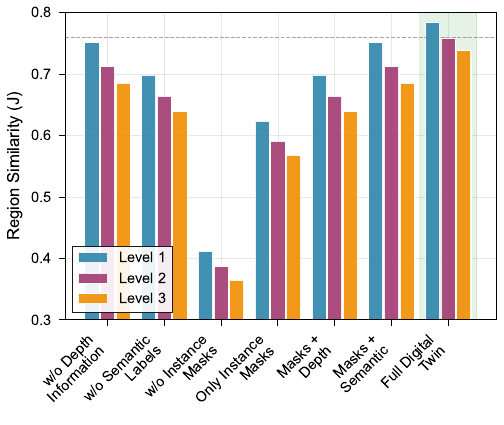}
\caption{Ablation study examining individual digital twin representation components on JiTBench~\cite{jit} in terms of region similarity ($\mathcal{J}$).}
\label{fig:digital_twin_ablation}
\end{figure}

\paragraph{Ablation Study}
Table~\ref{table:ablation} presents the ablation study, examining the contribution of each component in FastReasonSeg-1.7B-Distill at different difficulty levels in JiTBench~\cite{jit}. 
The digital twin representation proves important for reasoning segmentation performance, as replacing it with direct visual tokens as in most other baselines results in a dramatic performance decline from 0.760 to 0.588 average region similarity. 
The dynamic refinement capability through tool calling also contributes meaningfully, with static digital twin representations achieving only 0.699 average region similarity compared to the full model's 0.760.
The two-stage distillation design shows clear advantages over alternative training approaches.
Direct reinforcement learning without supervised fine-tuning achieves merely 0.669 average region similarity, while supervised fine-tuning alone reaches 0.716, confirming that the sequential combination of both stages produces superior results. 
%
%
Removing teacher guidance entirely by eliminating both reasoning and format rewards results in substantial performance degradation to 0.677 average region similarity.
%
%
Training the teacher model without reinforcement learning similarly impacts final performance, achieving 0.709 average region similarity.
Finally, Fig.~\ref{fig:digital_twin_ablation} demonstrates the individual contribution of each digital twin component to the overall performance.

\section{Conclusion}

We propose FastReasonSeg, a distillation framework for both image and video reasoning segmentation.
We innovatively decouple visual perception from reasoning via digital twin representations to enable compact models to perform reasoning segmentation without direct visual token processing. 
Moreover, our two-stage distillation preserves multi-step reasoning capabilities while reducing model parameters, demonstrating that reasoning quality need not be sacrificed for computational efficiency in reasoning segmentation. 
While the current implementation relies on pre-computed digital twin representations, FastReasonSeg opens pathways for real-time construction and refinement of these structured intermediates. 
%
%
%

{
    \small
    \bibliographystyle{ieeenat_fullname}
    \bibliography{main}
}


\end{document}